\documentclass{article}

\usepackage{PRIMEarxiv}
\usepackage{multirow}
\usepackage[utf8]{inputenc} 
\usepackage[T1]{fontenc}    
\usepackage{hyperref}       
\usepackage{url}            
\usepackage{booktabs}       
\usepackage{amsfonts}       
\usepackage{nicefrac}       
\usepackage{microtype}      
\usepackage{lipsum}
\usepackage{array}
\usepackage{fancyhdr}       
\usepackage{graphicx}       
\graphicspath{{media/}}     
\usepackage{bm}
\newcolumntype{C}[1]{>{\centering\arraybackslash}p{#1}}
\title{Check Field Detection Agent (CFD-Agent) using Multimodal Large Language and Vision Language Models
}

\author{
  Sourav Halder\thanks{Corresponding author.\\ \quad \textit{Email}: sourav.halder@usbank.com, halder.sourav92@gmail.com}\\
  U.S. Bank \\
  Chicago, IL, USA\\
   \And
  Jinjun Tong \\
  U.S. Bank \\
  Los Angeles, CA, USA\\
  \AND
  Xinyu Wu \\
  U.S. Bank \\
  Chicago, IL, USA\\
}

\begin{document}
\maketitle

\begin{abstract}
Checks remain a foundational instrument in the financial ecosystem, facilitating substantial transaction volumes across institutions. However, their continued use also renders them a persistent target for fraud, underscoring the importance of robust check fraud detection mechanisms. At the core of such systems lies the accurate identification and localization of critical fields, such as the signature, magnetic ink character recognition (MICR) line, courtesy amount, legal amount, payee, and payer, which are essential for subsequent verification against reference checks belonging to the same customer. This field-level detection is traditionally dependent on object detection models trained on large, diverse, and meticulously labeled datasets, a resource that is scarce due to proprietary and privacy concerns. In this paper, we introduce a novel, training-free framework for automated check field detection, leveraging the power of a vision language model (VLM) in conjunction with a multimodal large language model (MLLM). Our approach enables zero-shot detection of check components, significantly lowering the barrier to deployment in real-world financial settings. Quantitative evaluation of our model on a hand-curated dataset of 110 checks spanning multiple formats and layouts demonstrates strong performance and generalization capability. Furthermore, this framework can serve as a bootstrap mechanism for generating high-quality labeled datasets, enabling the development of specialized real-time object detection models tailored to institutional needs.
\end{abstract}


\section{Introduction}
Despite the growing adoption of electronic payment systems-including automated clearing house (ACH) transfers, wire transfers, and credit or debit cards-paper checks continue to play a critical role in financial transactions, particularly in high-value business-to-business (B2B) and government payments. The Federal Reserve reported approximately 36 million checks were processed in 2024 which amounted to a total of \$17.6 trillion \cite{ref2}. These figures underscore the continued economic significance of checks, even as digital payment alternatives proliferate.
Before the advent of AI, banks employed a combination of manual and simple rule-based systems to process and verify checks for fraud prevention (such as high amount or unusual locations). The rise of mobile check deposits via smartphone apps has improved convenience but also introduced new fraud risks due to poor image quality and digital tampering. These challenges highlight the need for robust, automated systems that can scale reliably. Advances in AI and machine learning now enable intelligent solutions that improve both the efficiency of mobile check processing and the detection of fraudulent activity.

A critical step in mobile check deposit and fraud detection is the accurate extraction of key fields from check images \cite{ref3, ref4}. These extracted features include but not limited to the signature, legal amount, courtesy amount, date, MICR line (routing number, account number, and check number), payer name, payee name, memo and bank name. These elements may carry signs of tampering and their reliable extraction is a crucial step in downstream tasks such as semantic analysis and anomaly detection. Once these check features are extracted, they can be further investigated for fraud detection such as comparing legal with courtesy amount \cite{ref3}, signature recognition \cite{ref5, ref6}, handwriting recognition \cite{ref7, ref8, ref9}, and so on. Machine learning (ML) based object detection models provide a promising pathway for automatic localization and extraction of critical check elements, enabling more scalable, consistent, and timely fraud prevention. 

Traditional ML-based object detection models for extracting check fields require fine-tuning on large, annotated datasets spanning diverse formats. However, curating such datasets is exceptionally challenging due to privacy constraints, limited availability, and the diverse visual structures of checks issued by different institutions. These challenges limit the scalability of supervised approaches. To address this, there is growing interest in training-free, generalizable frameworks that can detect check fields without labeled data or costly training. Recent advances in foundation models, such as large language models (LLM) and multimodal generative AI, show strong zero- and few-shot capabilities across tasks. Leveraging these models for financial document understanding could enable scalable, plug-and-play solutions that eliminate the need for task-specific training, accelerating intelligent automation in banking workflows. 

In this work, we present a framework that utilizes the combined capabilities of a vision language model (VLM) and a multimodal LLM (MLLM) to perform zero-shot object detection on bank checks. Although VLM is a multimodal language model, we consider it separately from MLLM since it lacks the enhanced language understanding and reasoning capability of MLLM.  The VLM predicts potential bounding boxes based on text prompts, and the MLLM identifies the correct bounding box from the VLM predicted bounding boxes. By harnessing the general image understanding and reasoning abilities of these models, our approach eliminates the need for task-specific training data or model fine-tuning. We evaluate this framework on a hand-curated dataset comprising 101 checks drawn from a variety of formats and issuing institutions. Despite the absence of any prior training on this dataset, the framework demonstrates strong performance in accurately identifying key check fields, underscoring its potential as a general-purpose solution for check image analysis. Furthermore, this system can serve as a data generation pipeline, automatically labeling check images to bootstrap high-quality datasets for training specialized object detection models, enabling faster inference and optimized performance for real-time deployment scenarios.

\section{Related Work}
\subsection{Object Detection Models}
Object detection has advanced from early CNN-based models like R-CNN \cite{ref10} and YOLO \cite{ref11} to transformer-based architectures such as DETR \cite{ref12} and Deformable DETR \cite{ref13}, which improve global context understanding. More recently, vision language models like OWL-ViT \cite{ref19}, OWLv2 \cite{ref20}, and Grounding DINO \cite{ref21} have enabled open-vocabulary and zero-shot detection using Vision Transformers \cite{ref22} and CLIP \cite{ref23}.These models are usually trained on large common object detection datasets such as MS COCO \cite{ref24} and they are focused towards a more generalized and flexible object detection paradigm. Although these object detection models have demonstrated good detection capabilities in a wide variety of applications, the biggest challenge in check field detection remains a challenge due to the lack of availability of large enough labeled datasets of checks for training these models. The creation of large and diverse datasets of bank check images, with a specific focus on fraudulent examples, is vital for training more accurate and generalizable models. 
\subsection{Multimodal Large Language Models (MLLM)}
Multimodal Large Language Models (MLLMs) extend traditional language models by integrating visual and textual inputs, enabling a richer, context-aware understanding. They perform well on image captioning and visual question answering, and are being explored for object detection, particularly in contextual and open-vocabulary settings. A promising avenue of active research is the application of MLLMs in object detection tasks. The current MLLMs, although successful in image captioning and visual QA, are not proficient at object localization tasks. Specifically, in the context of bank checks, MLLMs are highly accurate in text extraction tasks such as optical character recognition and named entity recognition, but not as accurate when it comes to detecting specific fields in checks.
 ContextDET \cite{ref26} advances contextual object detection by aligning visual objects with language input in interactive scenarios. Although effective, its reliance on task-specific training data limits its applicability, especially for domains like check field detection, where annotated datasets are scarce. Addressing another facet, the LLM-Optic \cite{ref27} method is introduced to resolve difficulties in visual grounding, particularly with intricate text queries. It capitalizes on the power of LLMs to bolster text query understanding using a Text Grounder, followed by using a pre-trained VLM to propose potential object locations in images and refining these proposals with a MLLM acting as a Visual Grounder. This enhancement allows sophisticated zero-shot visual grounding and shows remarkable improvements over existing benchmarks without the need for further model training. In this work, we extend the idea behind LLM-Optic to check field detection using agentic AI and OCR capabilities of MLLM. 

\section{Methodology}
\begin{figure}
\includegraphics[scale=0.5]{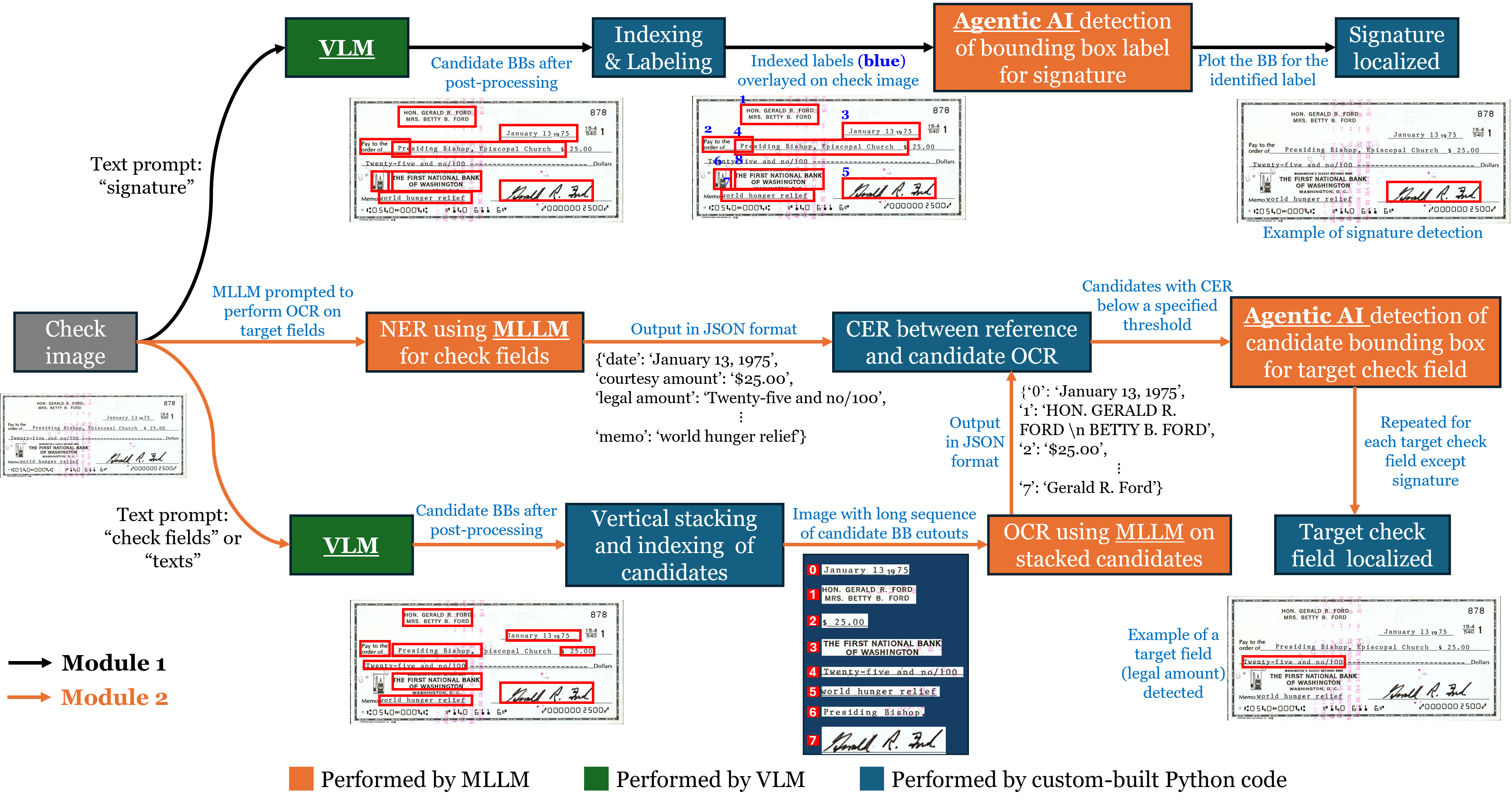}
\caption{Schematic of the CFD-Agent framework. The steps in module 1 and 2 are indicated by the black and orange arrows, respectively. The bounding boxes (BBs) shown in this figure are for illustrative purposes only.}
\label{fig:fig1}
\end{figure}
The CFD-Agent framework builds on the conceptual design of LLM-Optic with an additional enhancement using a multimodal agentic AI strategy for the domain-specific application of check field detection. An important step in the LLM-Optic framework is the use of an open vocabulary vision model, Grounding DINO, to perform object detection which resulted in candidate objects being selected. Although this strategy works well for open world object detection scenarios (since the VLM was trained on open world image data), resulting in only a limited number of candidate objects for the MLLM to select from, it has limited applicability in check field detection. This is because the named entity recognition (NER) task involved with check field detection is a highly specialized application and specific text prompts such as “payer name”, “signature”, etc., do not result in accurate detections. Rather, these prompts are interpreted by the VLM in a general manner such as “check fields” or “texts”, thus resulting in a huge number of detections of various texts and markings on checks, which are often overlapped. Therefore, a more enhanced and sophisticated framework is necessary to select the best bounding box from the large number of candidate boxes detected by the VLM. 

The schematic shown in Figure \ref{fig:fig1} describes the workflow of our framework as well as the intermediate outputs on a sample check. Our framework has two modules for check field detection based on the field of interest: module 1 for signature detection and module 2 for the detection of the remaining fields. This is because for all the check fields except signature, it is most often feasible to perform OCR, which is a crucial step in the detection process in module 2. Whereas, for the signature, performing accurate OCR is not a necessity and may not always be possible, and so follows a different approach in module 1. Based on our experiments, we observed that the approach in module 2 is faster and more accurate than that in module 1 for fields in which OCR can be performed, and therefore, the separation of the two modules was necessary for the best outcome.

\subsection{Vision Language Model (VLM)}
As described in the LLM-Optic article, VLMs do not have the capability to comprehend complex queries. So, the first step in that framework involves the use of an LLM to generate a short text prompt (usually one or two words) from a complex user query. In our case, we do not need that step since it involves the various field detections in checks which are always the same. So, the input prompt to the VLM is a predetermined text:
\begin{itemize}
    \item Module 1 text prompt: “signature”, 
    \item Module 2 text prompt: “check fields” for date, courtesy amount, legal amount, magnetic ink character recognition (MICR) line, memo; “texts” for payer name, payee name, bank name.
\end{itemize}

We chose a pre-trained OWLv2 \cite{ref20} as the VLM for CFD-Agent due to its versatile capability in zero-shot text-conditioned object detection which includes check field detection without any additional training requirement. We used the OWLv2 version with ViT-B/16 Transformer architecture \cite{ref22} on top of the CLIP \cite{ref23} backbone from the Hugging Face Transformers library. The check images were resized and padded to the shape 960×960×3 before to fit the input image dimension requirements for OWLv2. For detection, we used a confidence score threshold in the range 0.001-0.03 which resulted in detecting candidate bounding boxes for all the appropriate check fields. The main criterion for selecting this threshold was to ensure that important check fields were not missed in the detection process and at the same time the number of candidate detections was not overwhelmingly large (usually it was less than 80). By default, the maximum number of bounding boxes predicted is 3600. Non-max suppression was performed for post-processing to eliminate significantly overlapped detections with an Intersection-over-Union Threshold (IoU) specified as 0.4, which usually resulted in detections of candidate boxes in the range of 25-50. Furthermore, we also perform additional post-processing steps such as removing exceptionally large candidate bounding boxes (>25\% of the check image area) and candidate bounding boxes with extremely low dimensions (<12 pixels) or extremely high dimensions (>30\% of the corresponding check image dimension) to eliminate detection of different non-text markings such as lines.

\subsection{Module 1 (Signature Detection)}
\begin{figure}
\centering
\includegraphics[scale=0.7]{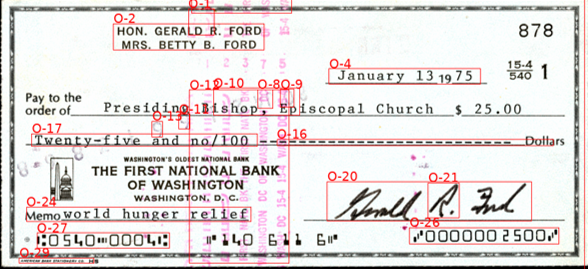}
\caption{Input to the MLLM (module 1) after labeling. The MLLM identifies the label for the target field, i.e., signature, which in this case should be O-20..}
\label{fig:fig2}
\end{figure}

Following the candidate bounding boxes detected by the VLM, as shown in Figure \ref{fig:fig2}, they are indexed, and the indexes are overlayed on the check images (labeling) according to their corresponding confidence scores. Note that even though “signature” is the text prompt for the VLM, it does not necessarily assign the highest confidence score for the candidate bounding box around the signature. This is because OWLv2 was trained on general open-world objects and not for specific tasks such as field detection in bank checks. The overlaid labels are used for the MLLM to use its general image understanding capabilities and select the label that corresponds to the signature field. As shown in Figure \ref{fig:fig2}, a downside to this approach is that the labels are often overlapped on top of each other, making it not feasible for the MLLM to identify the correct label. Thus, module 1 is used specifically for signature detection only. Note that this module can potentially be used for other non-text fields in checks as well such as the bank logo.
The detection of the signature in a bank check image is done using an agentic AI architecture composed of an Actor, an Environment, and an Evaluator. The system iteratively refines its selection based on feedback and memory of past errors. This conceptual framework was motivated by LLM-based verbal reinforcement learning \cite{ref28}. The main components in this framework are as follows:

\begin{itemize}
    \item Actor: A MLLM that receives a check image, candidate bounding boxes with labels, and a memory of past errors. It selects the label most likely to correspond to the target field (e.g., the payer signature),
    \item Environment: Visualizes the Actor’s selected bounding box on the check image and passes the rendered result to the Evaluator,
    \item Evaluator: Another MLLM that verifies whether the selected bounding box correctly identifies the target field. It returns a “Pass” with explanation if correct, or a “Fail” with feedback otherwise. If the grade assigned in “Fail”, then the selected candidate bounding box by the Actor is removed from the list of candidate bounding boxes for the next iteration,
    \item Memory: A dynamic list of misclassified labels. It evolves with each failed attempt and guides the Actor to avoid previously incorrect choices.
\end{itemize}
In order to formally define the algorithm, we introduce the following variables:
\begin{itemize}
    \item $\bm{\mathcal{J}}$: Bank check image,
    \item $n$: Total number of candidate bounding boxes after VLM and post-processing,
    \item $\bm{B}=[b_1, b_2, ...,b_n]$: A list of candidate bounding boxes produced by the upstream vision language model,
    \item $\bm{M}$: A memory set that stores prior misclassification labels. Initially the memory set is empty,
    \item $t$: Iteration step starting with 0,
    \item $T_{max}$: The maximum number of allowable iterations,
    \item $\bm{S}_t = \left(\bm{\mathcal{J}}, \bm{B}_t, \bm{L}_t, \bm{M}_t\right)$: Each state at iteration $t$.
\end{itemize}

\begin{figure}
\centering
\includegraphics[scale=0.8]{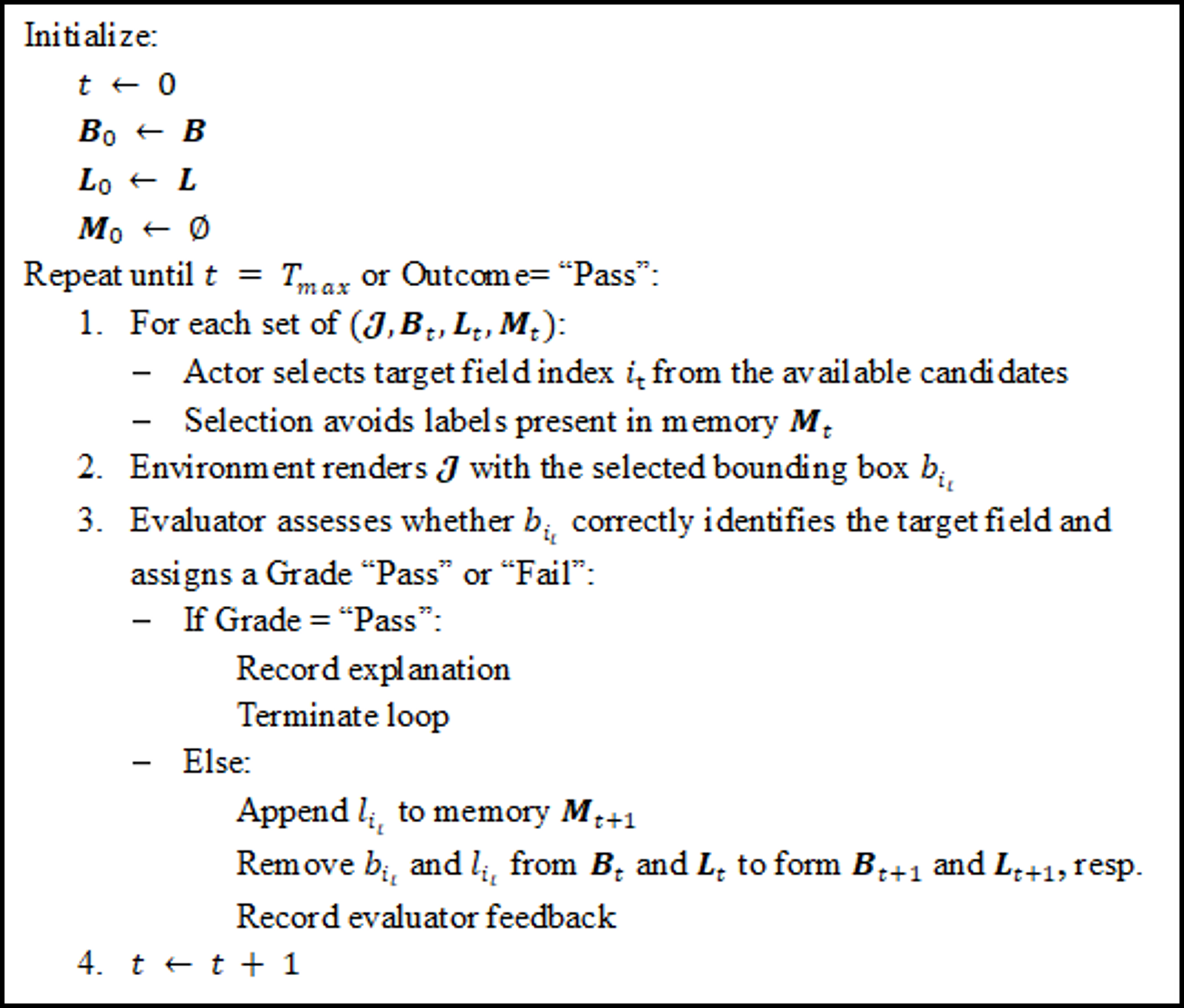}
\caption{Agentic AI algorithm for module 1.}
\label{fig:fig3}
\end{figure}
Figure \ref{fig:fig3} describes the agentic AI algorithm for selecting the correct bounding box from the list of candidates using the MLLM. In this case, we used GPT-4 as the MLLM, although the algorithm itself is agnostic to the type of MLLM used.

\subsection{Module 2 (OCR-based Check Field Detection)}
\label{section-module2}
Unlike the signature where an individual expresses their distinct style and may not always be accurately readable, all the other check fields are written for the purpose of proper readability. Thus, performing accurate OCR in these fields is generally possible. Module 2 utilizes Named Entity Recognition (NER) and OCR capabilities of MLLMs in addition to its general image understanding capabilities to implement a robust method for check field detection following the candidate bounding boxes predicted by the VLM. Just like module 1, it begins with a list of candidate bounding boxes. But instead of overlaying labels on the candidates, it stacks them vertically and indexes them using labels to their left as shown in Figure \ref{fig:fig1} This is done in order to avoid the overlap of labels as often observed in module 1. Following this, the MLLM is prompted to perform OCR on the vertically stacked fields along with the associated labels. Additionally, the MLLM is also prompted to perform NER of specific check fields given the original check image which becomes the reference. For each check field, character error rate (CER) is calculated between the reference for that field obtained by the MLLM-based NER and all the objects in the vertically stacked candidates. The CER is calculated based on the following formula:
\begin{equation}
C=\frac{S+D+I}{N},
\end{equation}
where $C$ is the CER, $S$ is number of character substitution, D is the number of character deletions, $I$ is the number of character insertions, and $N$ is the length of the reference text obtained via NER by MLLM. The CER was calculated using the opensource Python library editdistance \cite{ref29}, which is essentially a fast implementation of the Levenshtein distance. Based on a predefined threshold for CER ($C_o$), candidates from the vertical stack are selected with CER values less than $C_o$. Based on our experiments, we observed that specifying $C_o=0.8$ gave the best results overall. If more than one candidate is selected, then a similar agentic evaluation process is undergone as described for module 1, with the following modifications:
\begin{itemize}
    \item 	No separate MLLM actor is necessary anymore since the iteration is done over all the candidates that have CER values less than $C_o$,
    \item The set of bounding boxes $\mathbf{B}$ and the corresponding labels $\mathbf{L}$ do not contain all the candidates following the VLM and post-processing, but only a subset of size m which have CER less than $C_o$,
    \item The memory set is not required any more since no MLLM actor is needed.
\end{itemize}

\begin{figure}
\centering
\includegraphics[scale=0.8]{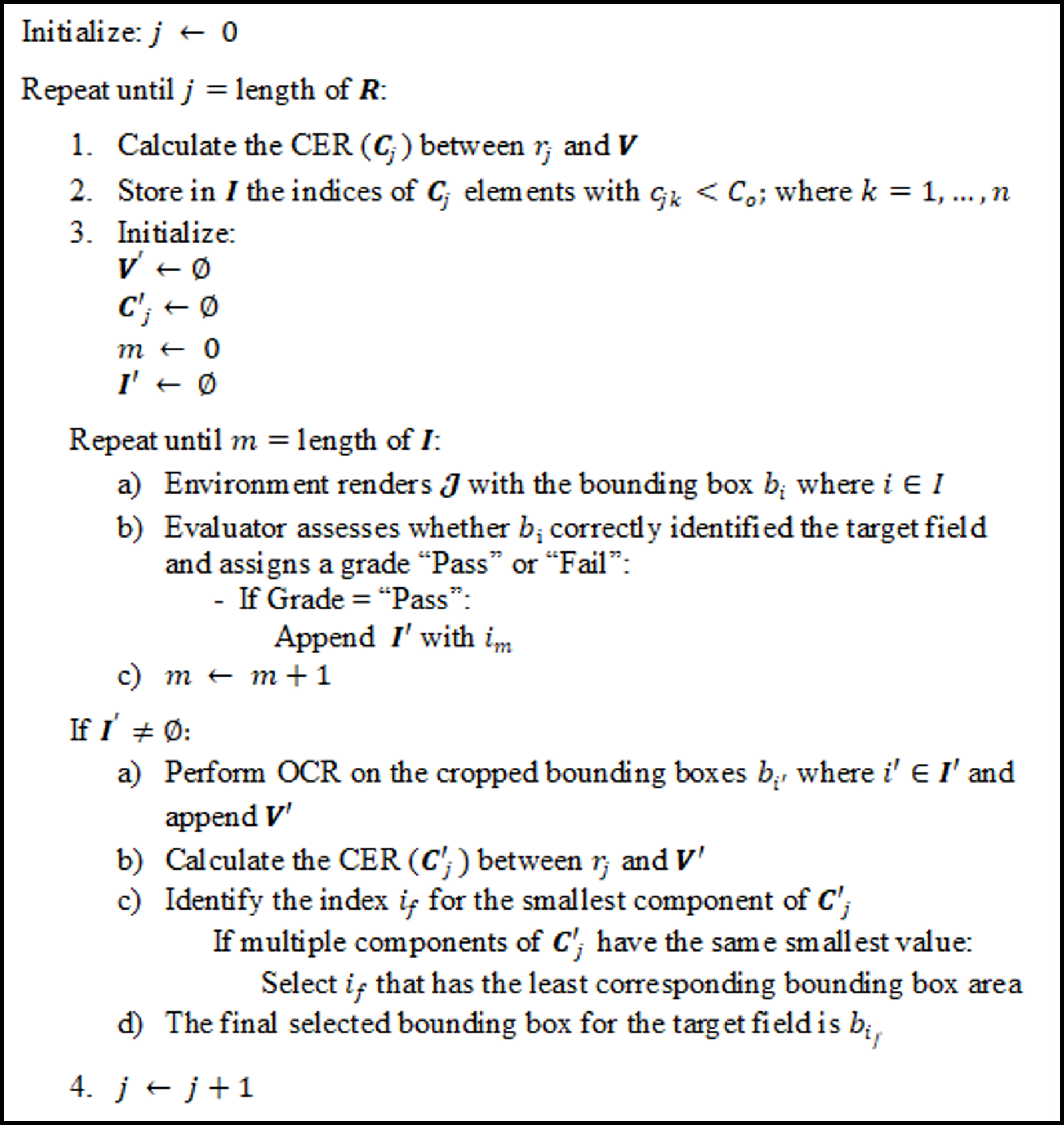}
\caption{Agentic AI algorithm for module 2.}
\label{fig:fig4}
\end{figure}

In addition to the variables defined for module 1, we introduce the following additional variables:
\begin{itemize}
    \item $\bm{V}=[v_1, v_2, ...,v_n]$: A list representing the texts from the vertical stack of candidate bounding boxes, where $n$ is the total number of candidate bounding boxes after VLM and post-processing,
    \item $\bm{R}=[r_1, r_2, ...,r_k]$: A list representing the reference texts from the NER conducted by the MLLM, where $k$ is the total number of check fields of interest. In this case, $k=8$ which includes all the check fields except the signature,
    \item $\bm{C}_j=[c_{j1}, c_{j2}, ...,c_{jn}]$: 	A list of CERs between each reference field text $r_j$ and all the texts extracted from the vertical stack of candidate bounding boxes, where $j=1,...,k$.
\end{itemize}

The algorithm for module 2 is shown in Figure \ref{fig:fig4}. Note that while stacking the candidates removes the problem of label overlapping, it also eliminates any positional information with respect to the original check image. Thus, the iterative MLLM based evaluation described above ensures that proper localization of the field of interest is executed. For instance, both the payer name and the signature might have the same CER with respect to the payer name reference, and the MLLM-based evaluation of the location of the candidate bounding box ensures that the correct candidate is selected. If more than one candidate is selected in the evaluation stage above, then the candidate bounding box with the smallest CER is selected. If there are more than one candidate bounding boxes with the smallest CER, then the one with the smallest bounding box area is selected.
\subsubsection{Additional Considerations and Notes}
A major challenge arises during the detection of the MICR line in the check images. First, the structure of the MICR line varies between checks of different banks. For instance, some checks have big gaps between the different components of the MICR line such as the routing number, account number and check number, while others do not. Thus, while detecting the various fields in check images, the VLM often detects only a part of the MICR line. Additionally, as described in more details in the following section, the NER performed by the MLLM is the least accurate for the MICR line compared to the other check fields. This adds to further inaccuracy in the detection of the MICR line. To tackle this problem, we only use the vertical coordinates of the bounding box for the MICR line detected by the framework and extend the horizontal coordinates to the ends of the check. For instance, if a check image has the dimension $960 \times 480$ (width  $\times$ height), and the coordinates of the bounding box were detected as $[200, 440, 820, 460]$ (in the form $[x_1,y_1, x_2, y_2]$ where $(x_1,y_1)$ and $(x_2, y_2)$ are the top-left and bottom-right coordinates, respectively), then the final post-processes bounding box coordinates would be specified as $[0, 440, 820, 960]$. 
Additionally, we have observed that if the vertical stack of candidate bounding boxes was too long, the accuracy of the MLLM in performing OCR decreases. To avoid that, we decided to split the vertical stack into multiple images with each limited to having up to 7 bounding boxes stacked. These smaller stacks are sent to the MLLM as input one at a time to perform OCR and later consolidated to maintain their original sequence before splitting.
Finally, note that although we used the MLLM to perform OCR on the vertical stacks of candidate bounding boxes, our framework conceptually allows for using other OCR models such as Tesseract \cite{ref30} and TrOCR \cite{ref31} to reduce the overall cost of running this framework while using commercial MLLMs such as GPT-4 or Claude 3.5. Our choice of using the same MLLM for NER to extract the reference texts and the OCR on the vertical stacks of candidate bounding boxes stems from minimizing the overall conceptual complexity of our framework by limiting the use of multiple different models and minimizing discrepancies in OCR while performing NER for reference texts and OCR on the vertical stacks of candidate bounding boxes.

\section{Experiments}
\subsection{Dataset}
The CFD-Agent framework performs zero-shot detection of bank check fields using the combined open world object detection capability of VLM and the general image understanding and OCR capability of MLLM. Thus, no training dataset was required for the CFD-Agent. To evaluate the efficacy of the CFD-Agent framework, a dataset comprising 110 bank check images was curated from publicly available sources on the internet due to the lack of publicly available datasets of bank checks for confidentiality. The dataset encompasses a diverse range of U.S. check formats, including personal checks, cashier’s checks, counterfeit specimens, and images exhibiting varying degrees of noise and degradation. This heterogeneity was deliberately introduced to simulate real-world conditions and ensure robustness across a spectrum of practical scenarios. The inclusion of both genuine and forged checks, along with variations in resolution, lighting, and occlusions, allows for a comprehensive assessment of the framework’s generalizability and resilience in the face of non-ideal inputs commonly encountered in financial document processing. We manually labeled all the check images with bounding boxes for each of the 9 check fields using the opensource VGG Image Annotator tool \cite{ref32} for the purpose of evaluation. We also manually extracted each field for NER evaluation. We used 5 check images from this dataset to develop the CFD-Agent framework, but none of the images were used to train any trainable parameters of either the VLM or the MLLM. 

\subsection{Evaluation}
The evaluation of the CFD-Agent framework falls under two categories: NER evaluation and evaluation of object detection for check fields.
\subsubsection{Evaluation of Named Entity Recognition (NER)}
The module 2 of the CFD-Agent framework heavily depends on accurate NER to extract the reference for the check fields. To evaluate the NER performance of GPT-4 (for module 2 of CFD-Agent), we manually extracted all the fields of interest which were used as ground truths. The CER was calculated between these ground truths and the GPT-4 performed NER using the ‘editdistance’ Python library as described in Section \ref{section-module2}. Finally, the statistical summary of the CER estimates is reported.  

\subsubsection{Evaluation of Object Detection}
Agent is to perform object detection on the various check fields. Some of the most widely used metrics for object detection are based on Intersection-over-Union (IOU) between ground truth bounding boxes and the model predicted bounding boxes. In this work, we used accuracy measures at 0.25 and 0.5 IOU thresholds. Note that other popular object detection evaluation metrics such as average precision (AP) and mean average precision (mAP) are not applicable for this framework. This is because these metrics require prediction of confidence scores by the object detection model. The design of the CFD-Agent framework (similar to LLM-Optic) does not have any confidence score associated with the final detected objects. Although the VLM predicts bounding boxes with associated confidence scores, the final bounding box selected by the MLLM rarely is the one with the highest confidence score. Thus, these confidence scores are not relevant for estimating mAP. Therefore, we performed benchmarking tests on our CFD-Agent framework against LLM-Optic, which is the only comparable general purpose visual grounding model available in the literature (to the best of our knowledge) that uses MLLM for object detection without requiring any training or fine-tuning. The metrics used for this benchmarking includes mean intersection over union (mIOU), accuracy at IOU thresholds of 0.25 and 0.5. This evaluation method includes the predictions of both module 1 and 2 of CFD-Agent.

\subsection{Results}
\subsubsection{Named Entity Recognition (NER)}
Table \ref{tab:table1} describes the character error rates (CER) between the MLLM performed NER and manually labeled ground truth for 8 check fields. The CER results reveal several key trends in the performance of GPT-4 as the MLLM in the NER task across the different fields in bank checks.  Note that since CER is an error metric, lower values indicate better performance. Overall, GPT-4 achieves strong accuracy in many fields, with an overall weighted average CER of 0.070 across all the 8 check fields. 
\begin{table}
 \caption{Character error rate (CER) for MLLM-based NER using human evaluation}
  \centering
  \begin{tabular}
    {p{0.2\textwidth}||C{0.1\textwidth}C{0.1\textwidth}C{0.1\textwidth}C{0.05\textwidth}}
    \toprule
    Check field     & Mean     & Std     &Median     &Total \\
    \midrule
    Date & 0.049  & 0.113  &0.000   &108     \\
    Courtesy amount	&0.051	&0.127	&0.000	&107 \\
    Legal amount	&0.061	&0.117	&0.000	&99 \\
Payer name	&0.111	&0.435	&0.000	&88 \\
Bank name	&0.025	&0.164	&0.000	&93 \\ 
Memo	&0.014	&0.064	&0.000	&56 \\
MICR	&0.134	&0.137	&0.104	&95 \\ 
Payee name	&0.083	&0.450	&0.000	&95 \\
    \bottomrule
  \end{tabular}
  \label{tab:table1}
\end{table}

The mean CER values demonstrate high accuracy with the mean CER less than 10\% for most of the fields except MICR and payer name. However, the high standard deviations observed in the fields payee name (0.450) and payer name (0.435) point to a larger deviation from their corresponding mean CER values that dominate the overall error rates. The sample sizes of all the check fields are not the same, which may have some impact on the relative differences in the CER aggregates. This is because of the wide variety of checks in the dataset with some missing fields. These include cashier’s checks, checks for instructional purposes with some fields removed, checks with illegible handwriting among others. Memo is notable among the other check fields since it is often an optional field and was missing in many of the checks.

Overall, long check fields such as the MICR field and handwritten fields such as names and legal amounts remain the most error-prone. The latter are typically associated with challenging handwriting, rare or foreign names, or poor image quality, which cause NER to fail completely in some samples. Somewhat surprising result was the relatively high CER in the MICR field, with a mean of 0.134 and a median of 0.104. MICR fonts are standardized and machine-readable by design, so performance degradation in this field likely stems from the inherent limitations of the MLLM (GPT-4) is performing OCR in long fields. Our limited experiments on various prompting strategies did not result in significant improvements in the NER results for MICR. An extensive investigation into prompt engineering could potentially improve this performance. Additionally, a wide-scoped experimentation with different MLLM could potentially result in the selection of the best MLLM for the highest NER accuracy, but that is beyond the scope of this work. 

The overall performance of the CFD-Agent, specifically the module 2 in its framework, is highly dependent on the successful implementation of the NER. As described in Figure \ref{fig:fig4}, module 2 allows for some inaccuracies in the NER. As long as the CER threshold $C_o$ is high enough to allow for minor errors in the CER, the agentic reasoning feature in module 2 still makes it possible to detect the correct fields as prompted. In this work we specified $C_o=0.75$, which is high enough to incorporate any CER as shown in Table \ref{tab:table2}. Additionally, since the same MLLM was used in NER as the reference text for each field and to perform OCR on the vertically stacked candidate objects, similar errors would potentially be made in both these tasks resulting in low relative CER between them.

\subsubsection{Check Field Detection}
\begin{figure}
\centering
\includegraphics[scale=0.6]{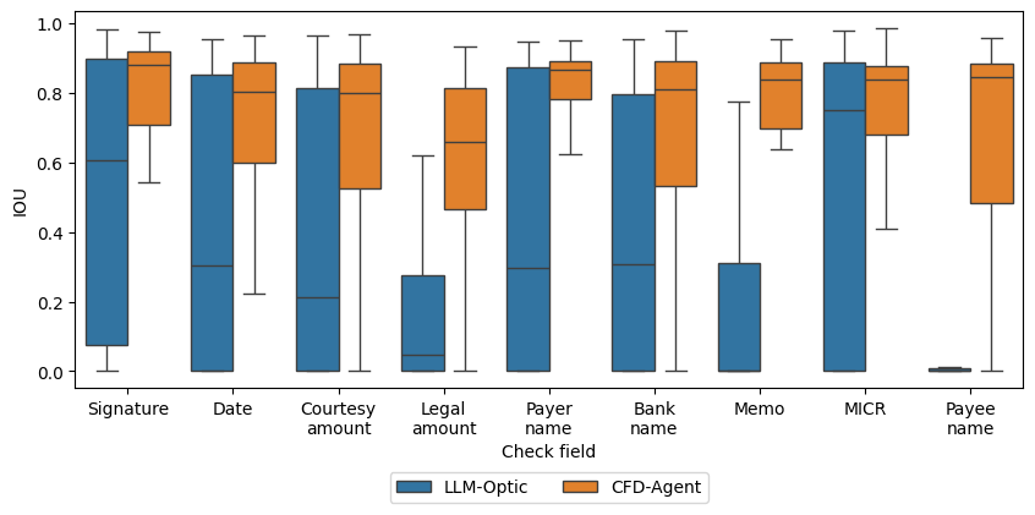}
\caption{Benchmarking of CFD-Agent with LLM-Optic.}
\label{fig:fig5}
\end{figure}

The accuracy of the check field detection was estimated using Intersection-over-Union (IOU). Figure \ref{fig:fig5} plots the IOU distributions for the nine target fields (Signature, Date, Courtesy amount, Legal amount, Payer name, Bank name, Memo, MICR, and Payee name). CFD-Agent consistently achieves higher or more stable IOU values across all the fields compared to LLM-Optic. This improvement is particularly noticeable in challenging fields such as “Legal amount,” “Payer name,” and “MICR,” where LLM-Optic’s predictions display a wide variance and lower median IOU values. In contrast, CFD-Agent’s predictions in these categories are both tighter in distribution and exhibit higher medians, indicating more reliable localization. In cases like “Legal amount”, “Memo” and “Payee name,” LLM-Optic struggles considerably with very low median IOU values against the corresponding ground truths, while CFD-Agent maintains a robust baseline performance.

\begin{table}
 \caption{CFD-Agent performance benchmark on 110 checks}
  \centering
  \begin{tabular}{p{0.2\textwidth}||ccc|ccc}
    \toprule
    \multirow{2}{*}{Check field}& \multicolumn{3}{c|}{CFD-Agent} & \multicolumn{3}{c}{LLM-Optic} \\
    \cmidrule(r){2-7}
        & mIOU	&Acc@0.25	&Acc@0.5	&mIOU	&Acc@0.25	&Acc@0.5 \\
    \midrule
    Signature & 0.692	&0.781	&0.781	&0.529	&0.619	&0.562 \\
    Date	&0.691	&0.870	&0.815	&0.417	&0.546	&0.444 \\
Courtesy amount	&0.680	&0.897	&0.766	&0.372	&0.486	&0.421 \\
Legal amount	&0.628	&0.939	&0.697	&0.186	&0.283	&0.121 \\
Payer name	&0.761	&0.909	&0.852	&0.430	&0.523	&0.466 \\
Bank name	&0.680	&0.871	&0.774	&0.373	&0.516	&0.376 \\
Memo	&0.749	&0.929	&0.893	&0.221	&0.286	&0.232 \\
MICR	&0.750	&0.979	&0.863	&0.519	&0.642	&0.558 \\ 
Payee name	&0.680	&0.863	&0.768	&0.120	&0.168	&0.105 \\ 
Overall mean	&0.698	&0.796	&0.890	&0.360	&0.374	&0.462 \\

    \bottomrule
  \end{tabular}
  \label{tab:table2}
\end{table}

Table \ref{tab:table2} specifies CFD-Agent’s performance in comparison with LLM-Optic in terms of three object detection metrics. CFD-Agent substantially outperforms LLM-Optic across all evaluation metrics. It achieves an overall mIOU of 0.698 compared to 0.360 for LLM-Optic, indicating significantly better localization quality. Accuracy at IOU thresholds further highlights this gap: CFD-Agent reaches 89.0\% at Acc@0.25 and 79.6\% at Acc@0.5, while LLM-Optic trails at 46.2\% and 37.4\%, respectively. These results confirm that CFD-Agent is both more accurate and more precise, especially excelling at stricter IOU thresholds like Acc@0.5, which demands more precise bounding boxes. This suggests that CFD-Agent is not only able to detect the presence of check fields but also localize them with greater precision demonstrating the effectiveness of agentic reasoning in tasks that require both visual understanding and contextual alignment. 

In Table \ref{tab:table1}, the MICR field had the worst accuracy for NER (highest mean CER), but as shown in Figure \ref{fig:fig5} and Table \ref{tab:table2}, the object detection accuracy is quite high. In fact, the Acc@0.25 was highest for MICR among other check fields. In comparison to LLM-Optic, the variation of the IOU (as shown in Figure \ref{fig:fig5}) was also significantly lower. This is because the CER threshold specified in module 2 of CFD-Agent was high enough to tolerate any errors in the OCR of the MICR field. The high interquartile range for IOU of LLM-Optic was due to the MLLM selecting other features such as horizontal lines or other patterns within the check. This is most likely due to the similarities in image encodings of the long sequence of digits in the MICR field and the pixel values of other patterns within the check. The strict requirement of module 2 of CFD-Agent using CER (instead of relying on the MLLM’s identification of the correct label as in module 1) for identifying relevant objects (or fields) within the checks ensures accurate localization of the MICR field.
The object detection of the legal amount by CFD-Agent had the least accuracy as shown by the mIOU and Acc@0.5 metrics in Table \ref{tab:table2}. This is because this field is one of the longest handwritten fields in bank checks. The VLM which predicts the candidate bounding boxes sometimes do not have a bounding box that completely encapsulates the legal amount field. This is evident from the observation that the legal amount localization gets a big lift of ~24 pp from Acc@0.5 to Acc@0.25. 

The biggest lift in accuracy of check field detection when compared with LLM-Optic as a baseline was observed for the payee name. Even with the most lenient accuracy metric, Acc@0.25, there was a ~70 pp improvement by CFD-Agent when compared with LLM-Optic. This is primarily due to a lot of overlap of the labels of various candidate bounding boxes predicted by the VLM near the vicinity of the payee name. Module 2 of CFD-Agent significantly mitigates this problem.

Overall, the results underscore a critical insight into the application of VLMs and MLLMs in structured financial documents: general-purpose frameworks like LLM-Optic, while capable in open-domain scenarios, lack the inductive biases and contextual rigor required for precise field localization in bank checks. CFD-Agent’s performance highlights the value of incorporating agentic reasoning, where downstream task objectives such as OCR quality and semantic field constraints can dynamically guide localization decisions. Notably, its framework design enables refinement strategies, such as the use of CER-informed filtering, to disambiguate visually similar regions (e.g., in the MICR field) and resolve overlapping candidate predictions (e.g., for the payee name). CFD-Agent’s superior performance in stricter metrics like Acc@0.5 also signals its strength in precisely delineating their spatial boundaries which is crucial for various downstream tasks such as check fraud detection and information extraction. Importantly, the observed limitations, such as partial capture of long handwritten fields like the legal amount, suggest that bounding box generation by the VLM still poses a bottleneck and may benefit from improved object detection models specialized in document analysis.

While this study demonstrates the efficacy of CFD-Agent using a specific combination of component models: GPT-4 as the MLLM and OWLv2 as the VLM, the framework itself is modular and not inherently tied to these choices. Other MLLMs and VLMs can be substituted within the agentic loops, potentially offering better performance depending on the model characteristics and the nature of the document set. However, a comprehensive investigation into the optimal configuration of these component models is beyond the scope of this work. Nonetheless, the flexibility of CFD-Agent opens up promising avenues for further research into model composition strategies for document AI in high-stakes, domain-specific settings.

\section{Conclusion}
In this work, we described an agentic AI framework called CFD-Agent that uses a vision language model (VLM) and multimodal LLM (MLLM) to perform zero-shot check field detection on nine fields: signature, date, courtesy amount, legal amount, payer name, bank name, memo, MICR, and payee name. The CFD-Agent framework has two modules depending on the check field to be detected: module 1 for signature and module 2 for the rest. Both modules use the VLM (OWLv2) to predict the initial candidate bounding boxes, followed by using the general image understanding, OCR, and reasoning capabilities of MLLM to identify the correct bounding box for each check field from among the candidate bounding boxes. The performance of the CFD-Agent framework was evaluated using a dataset of 110 check images using character error rates for NER and mean intersection over union (IOU), accuracy at 0.25 and 0.5 IOU thresholds for object detection. CFD-Agent was benchmarked against a comparable general purpose object detection algorithm and demonstrated superior performance across all nine check fields on all the metrics. Since CFD-Agent is highly accurate and does not require procuring a large dataset for any training, which is a challenging task due to the highly confidential nature of financial documents and lack of public availability of such datasets, it is well-suited for adaptation and integration into a wide variety of downstream systems involving check fraud detection, automatic reconciliation, and document-based KYC pipelines.

\bibliographystyle{unsrt}  
\bibliography{cfd_agent}

\end{document}